\documentclass{article}

\usepackage{microtype}
\usepackage{graphicx}
\usepackage{subcaption}
\usepackage{booktabs} % for professional tables
\usepackage{amssymb}
\usepackage{amsmath}
\usepackage{balance}
\usepackage{wrapfig}
\usepackage{tikz}
\usepackage{pgfplots}
\usepackage{math-style}
\usepackage{enumitem}
\usepackage{hyperref}

\usepackage[accepted]{icml2019}

\icmltitlerunning{Supervized Segmentation}

\begin{document}

\twocolumn[
\icmltitle{Supervized Segmentation with Graph-Structured Deep Metric Learning}




\begin{icmlauthorlist}
\icmlauthor{Loic Landrieu\textsuperscript{$\star$}}{ign}
\icmlauthor{Mohamed Boussaha\textsuperscript{$\dagger$}}{ign}
\end{icmlauthorlist}

\icmlaffiliation{ign}{Univ. Paris-Est, IGN-ENSG, LaSTIG, \textsuperscript{$\star$}STRUDEL, 
\textsuperscript{$\dagger$}ACTE, Saint-Mand\'e, France}

\icmlcorrespondingauthor{Loic Landrieu}{loic.landrieu@ign.fr}
\icmlkeywords{Metric Learning, Graph, point clouds, segmentation}

\vskip 0.3in
]
%\setlength{\belowcaptionskip}{-10pt}
%\addtolength{\textfloatsep}{-5mm}
%\setlength{\abovedisplayskip}{2mm}
%\setlength{\belowdisplayskip}{2mm}

\printAffiliationsAndNotice{ } % otherwise use the standard text.

\begin{abstract}
We present a fully-supervized method for learning to segment data structured by an adjacency graph. 
We introduce the \emph{graph-structured contrastive loss}, a loss function structured by a ground truth segmentation. It promotes learning vertex embeddings which are homogeneous within desired segments, and have high contrast at their interface. Thus, computing a piecewise-constant approximation of such embeddings produces a graph-partition close to the objective segmentation.
This loss is fully backpropagable, which allows us to learn vertex embeddings with deep learning algorithms.
We evaluate our methods on a 3D point cloud oversegmentation task, defining a new state-of-the-art by a large margin.
These results are based on the published work of \citet{landrieu2019supervized}.
\end{abstract}
\vspace{-5mm}
\section{Introduction}
We consider the problem of learning to segment data points into meaningful groups. More precisely, we consider the case in which such points are linked by a \emph{sparse} graph-structure, and each point is attributed with expressive features. In this case, segmentation can be viewed as a graph partitioning problem based on learned vertex embeddings. 

The task of segmentation has been extensively studied for images \citep{achanta2012slic}, 3D point clouds \citep{LIN201839, PaponASW13}, and community retrieval \citep{fortunato2010community}. A common roadblock is that segmentation operators are not backpropagable. The first problem is that the codomain of such operators is the set of vertex partitions for which no simple metric can be used to compute derivatives. Furthermore, graph partitions generally rely on computing connected components, which is highly discontinuous as a single edge can completely overhaul the partition. These limitations prevent directly using segmentation metrics for learning point embeddings. 

These issues are addressed by \citet{JampaniSLYK18} for images by using \emph{soft-affectations}, which render the segmentation process continuous and allow for backpropagation. However, these improvements come with strong assumptions on the uniform size and distribution of the segments. While these are reasonable for computing superpixels, our objective is to allow for adaptive segment size and density in order to deal with other types of data.

\citet{liu2018learning} proposes a new loss function, called the \emph{segmentation-aware loss} (SEAL), for learning pixel embeddings while taking into account their influence on the quality of the segmentation. In this paper, we propose an improved way of incorporating guidance from segmentation, which is simultaneously more stable and accurate.

Community discovery methods are largely based on graph topology analysis rather than point embeddings, and hence are beyond the scope of this paper. Likewise, the random forest-based supervised graph partitioning algorithm of \citet{reas2018superpart} aims at learning to recognize transitions between segments rather than learning vertex embeddings. Finally, the work of \citet{liao2018graph} uses graph partitioning to help classification rather than learn to segment.
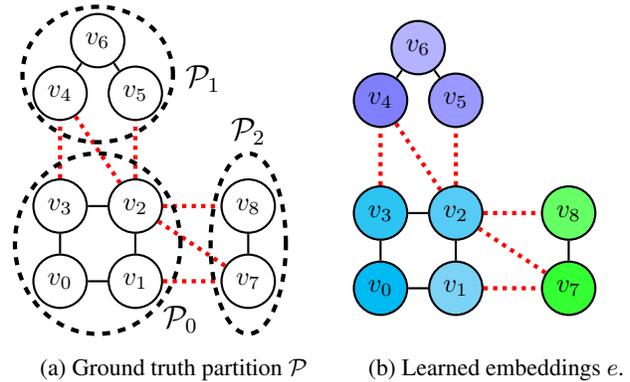
\begin{figure}
\setlength\tabcolsep{0pt}
    %auto-ignore
\begin{tabular}{cc}
 \begin{subfigure}[b]{0.25\textwidth}
 \begin{tikzpicture}
    \tikzstyle{vertex}=[circle, draw = black, thick]
    \tikzstyle{trans}=[ultra thick, draw = red, dotted]
    \tikzstyle{edge}=[thick, draw = black]
        \node at (0,0) [vertex,fill = white] (v0){$v_0$};
        \node at (1,0) [vertex,fill = white] (v1){$v_1$};
        \node at (1,1) [vertex,fill = white] (v2){$v_2$};
        \node at (0,1) [vertex,fill = white] (v3){$v_3$};
        
        \node at (0,2.5) [vertex,fill = white] (v4){$v_4$};
        \node at (1,2.5) [vertex,fill = white] (v5){$v_5$};
        \node at (0.5,3.2) [vertex,fill = white] (v6){$v_6$};
        
        \node at (2.5,0) [vertex,fill = white] (v7){$v_7$};
        \node at (2.5,1) [vertex,fill = white] (v8){$v_8$};
        
        \draw [edge] (v0) -- (v1) -- (v2) -- (v3) -- (v0);
        \draw [edge] (v8) -- (v7);
        \draw [edge] (v4) -- (v6) -- (v5);
        \draw [trans] (v1) -- (v7) -- (v2) -- (v8);
        \draw [trans] (v3) -- (v4) -- (v2) -- (v5);
        
        \draw [dashed, ultra thick] (0.5,0.5) ellipse (1.1cm and 1.2cm);
        \draw [dashed, ultra thick] (0.5,2.75) ellipse (1cm and .9cm);
        \draw [dashed, ultra thick] (2.5,0.5) ellipse (0.5cm and 1.2cm);
        
        \node at (1.6,-0.5) [draw = none] (l0){\large $\cP_0$};
        \node at (1.9,2.7) [draw = none] (l1){\large $\cP_1$};
        \node at (2.5,2) [draw = none] (l2){\large  $\cP_2$};
    \end{tikzpicture}
    \caption{Ground truth partition $\cP$}
    \label{fig:gt}
    \end{subfigure}
    &
    \begin{subfigure}[b]{0.25\textwidth}
      \begin{tikzpicture}
    \tikzstyle{vertex}=[circle, draw = black, thick]
       \tikzstyle{trans}=[ultra thick, draw = red, dotted]
    \tikzstyle{edge}=[thick, draw = black]
        \node at (0,0) [vertex,fill = cyan!80!white] (e0){$v_0$};
        \node at (1,0) [vertex,fill = cyan!45!white] (e1){$v_1$};
        \node at (1,1) [vertex,fill = cyan!55!white] (e2){$v_2$};
        \node at (0,1) [vertex,fill = cyan!65!white] (e3){$v_3$};
        
        \node at (0,2.5) [vertex,fill = blue!50!white] (e4){$v_4$};
        \node at (1,2.5) [vertex,fill = blue!40!white] (e5){$v_5$};
        \node at (0.5,3.2) [vertex,fill = blue!30!white] (e6){$v_6$};
        
        \node at (2.5,0) [vertex,fill = green!80!white] (e7){$v_7$};
        \node at (2.5,1) [vertex,fill = green!60!white] (e8){$v_8$};
        
        \draw [edge] (v0) -- (v1) -- (v2) -- (v3) -- (v0);
        \draw [edge] (v8) -- (v7);
        \draw [edge] (v4) -- (v6) -- (v5);
        \draw [trans] (v1) -- (v7) -- (v2) -- (v8);
        \draw [trans] (v3) -- (v4) -- (v2) -- (v5);
        
        \draw [draw = none] (0.5,0.5) ellipse (1.1cm and 1.2cm);
        \draw [draw = none] (0.5,2.75) ellipse (1cm and .9cm);
        \draw [draw = none] (2.5,0.5) ellipse (0.5cm and 1.2cm);
    \end{tikzpicture}
        \caption{Learned embeddings $e$.}
        \label{fig:emb}
   \end{subfigure}
    \end{tabular}
    \caption{Illustration of our method: vertex embeddings \Subref{fig:emb} are learned such that they are homogeneous within ground truth segments $\cP$ \Subref{fig:gt}, and with high contrast at transition edges (in red).}
    \label{fig:intro}
\end{figure}
\vspace{-3mm}
\section{Method}
We propose a two step approach to graph partitioning supervized by a ground truth segmentation. 
First, we compute vertex embeddings which are homogeneous within segments and present high contrasts at their border; second, we compute a piecewise-constant approximation of these embeddings with respect to an adjacency graph, as represented in \figref{fig:intro}.
As suggested by \citet{wang2014learning}, we bound the embeddings to the unit sphere to prevent their collapse during learning.

We consider a set $V$ of data points, whose adjacency structure is encoded by the graph $G=(V,E)$, with $E \subset V\times V$ the set of edges. Throughout this paper, we will assume that this adjacency structure is \emph{sparse}, in the sense that $\Abs{E}\ll \Abs{V}^2$. Note that this is not necessary for the mathematical derivations of the method. However, the proposed method would not be well-suited for a segmentation problem with a complete adjacency structure.

For a partition $\cU=(U_1, \cdots, U_K)$ of  $V$, we denote $\Etra(\cU)$ its set of transition edges, \ie linking different elements of $\mathcal{U}$: $\Etra(\cU)=\{(u,v) \in E \mid u \in U_i, v \in U_j, i \neq j\}$.

\textbf{Generalized Minimal Partition.} We associate to each vertex $v$ an embedding $e_v$ in the $m$-dimensional unit sphere $\bbS_{m}=\{x \in \bbR^m \mid \Norm{x}=1\}$. 
For such embeddings, we can partition $V$ into the constant connected component of a piecewise-constant approximation of $e$ with respect to graph $G$. Such approximation can be defined as the solution of the Generalized Minimal Partition Problem (GMPP) introduced by \citet{landrieu2017cut}:
\begin{align}
\!\!\!\!e^\star & \in \argmin_{f \in \bbS_m^{V}}
\sum_{i\in V}\Vert f_i - e_i \Vert^2 +
\sum_{(i,j)\in E}w_{i,j} \Bra{f_i \neq f_j}~,
\label{eq:mgp}
\end{align}
with $w \in \bbR_+^{E}$ the edges' weight and $\Bra{ x \neq y}$ the Iverson's bracket equal to $0$ if $x=y$ and $1$ otherwise. To encourage splitting along high contrast areas, we define the edge weight as
$
w_{i,j} = \lambda \exp \Pa{\frac{-1}{\sigma} \Vert e_i - e_j \Vert^2},
$
with parameters $\lambda,\sigma\in\bbR^+$. This problem is noncontinuous, nondifferentiable, and nonconvex, and hence hard to solve. However, good approximate solutions can be efficiently computed with the $\ell\!-\!0$ cut-pursuit algorithm of \citet{landrieu2017cut}.
Thus, a given embedding $e$ defines a GMPP,
whose approximate solution is $e^\star$, whose constant connected components defines a segmentation, which we denote $\GMPP{e}$.

\textbf{Undersegmentation Error.} Given a proposed partition $\mathcal{S}=\{S_1, \cdots, S_L\}$ of $G$, one can define its agreement with a ground truth segmentation $\cP$ through the \emph{undersegmentation error} $\mathcal{L}$ \citep{levinshtein2009turbopixels}, which sums over each segment $S_l$ the number of vertices which are not in the \emph{majority true segment}, \ie the element of $\cP$ with the largest overlap with $S_l$:
$$
\mathcal{L}(P,S) = \frac1{\Abs{V}}\sum_{l=1}^L \min_{P \in \cP} \Abs{ S_l \setminus P}~.
$$
\textbf{Learning Embeddings for Segmentation.}
Our objective is to learn a vertex embedding function $\xi:V \mapsto \bbS_{m}$ such that $\xi(V)$ is homogeneous within the segments of the ground truth segmentation $\cP$, and with high contrast at transition edges $\Etra(\cP)$. This property encourages $\Etra(\GMPP{\xi(V)})$---the transition edges between constant connected components of the piecewise approximation of $e$ --- to be close to $\Etra(\cP)$. The function $\xi$ is typically a neural network operating on features of the data points corresponding to the vertices of $V$. This can be for example the color of the pixels of an image, or the local geometry/radiometry of 3D points in a point cloud. Furthermore, these features can be computed from the neighbors of each vertex in $G$, allowing for the use of a wide range of networks such as convolution-based architectures.

\textbf{Graph-Structured Contrastive Loss.}
The naive way to learn such an embedding function $\xi$ would be to minimize the undersegmentation error $\cL(\GMPP{\xi(V)},\cP)$ directly. However, because the optimization problem defined in \eqref{eq:mgp} is noncontinuous and nonconvex, it is difficult, if not impossible, to backpropagate through its minimization. As discussed earlier, the constant connected component operator is not backpropagable either. Furthermore, the undersegmentation error would favor very granular partitions as it doesn't penalize high segment counts.

Consequently, we introduce $\ell$ the graph-structured contrastive loss, a surrogate loss function to the undersegmentation error, which operates on edges instead vertices and  allows for backpropagation:
\begin{align}\nonumber
\!\ell(e,\cP)\cdot{\vert E\vert}
\;\;= 
\!\!\!\!\!\!\!\!\!\!\!\!
\sum_{(u,v)\in E\setminus \Etra(\cP)}\!\!\!\!\!\!\!\!\!\!\!\!\phi\Pa{e_u - e_v}+
\!\!\!\!\!\!\!\!\!\!\!\sum_{(u,v)\in \Etra(\cP)}\!\!\!\!\!\!\!\!\!\!\! \mu_{u,v}^{(e)} \psi\Pa{e_u - e_v},
%}\!,
%\label{eq:gtl}
\end{align}
with $\phi$ (resp. $\psi$) a function favoring similarity (resp. contrast), and $\mu_{i,j}^{(e)} \in \bbR^{\Etra}$ a weight on transition edges, discussed later. A vertex embedding function minimizing this loss will be uniform within elements of $\cP$ and have high contrasts at $\Etra(\cP)$. Consequently, $\Etra(\GMPP{e})$ should be close to $\Etra(\cP)$.
%--------------------------------------------------------------------
\begin{wrapfigure}{R}{0.23\textwidth}
\resizebox{0.23\textwidth}{!}{
\begin{tikzpicture}
\begin{axis}[ xmin = -2, xmax = 2, ymin = 0, ymax = 2, xlabel=\huge $x$, ylabel={\huge\textcolor{blue}{$\phi(x)$}\;and\;\textcolor{red}{$\psi(x)$}},axis line style = ultra thick,axis lines=middle]
\addplot[blue,line width = 2pt,mark=none,samples=200] {0.3 * ((((x/0.3)^2 + 1)^0.5)-1)};
\addplot[color=red,mark=none,line width = 2pt] coordinates { (-2,0) (-1,0) (0,1) (1,0) (2,0)};
\end{axis}
\end{tikzpicture}
}
\caption{The functions $\phi$ (in blue) and $\psi$ (in red) used in the graph-structured contrastive loss.}
\label{fig:huber}
\end{wrapfigure}
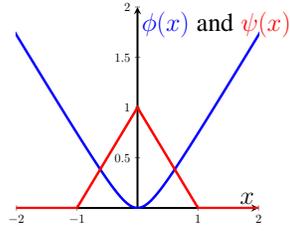
%--------------------------------------------------------------------
Our proposed loss is related to the contrastive loss of \citet{chopra2005learning} and the triplet loss popularized by both \citet{hoffer2015deep} and \citet{wang2014learning}. However, our method takes advantage of the adjacency structure.
%This loss differs from the , as it involves all vertices within a graph (or a sub-graph) at once, and not  an anchor and related positive/negative examples.
This allows us to bypass the problem of \emph{example picking} altogether. Indeed, the positive and negative examples are directly determined by the graph structure, instead of computationally intensive hard example mining. 

We chose $\phi$---the function promoting intra-segment homogeneity---as $\phi(x)=\delta (\sqrt{\Vert x \Vert^2 / \delta^2+1}-1)$ (represented in \figref{fig:huber}). This means that the first term of $\ell$ is the (pseudo)-Huber graph-total variation on the non-transition edges \citep{huber1973robust, charbonnier1997deterministic}, promoting smooth homogeneity of embeddings within segments.

With $\psi(x)=\max\Pa{1-\Vert x \Vert ,0}$, the second part of $\ell$ is the opposite of the truncated graph-total variation \citep{zhang2009some} on the transition edges. It penalizes similar embeddings at the edges between ground truth segments. As the embeddings are spherically-bound, we threshold this function for differences larger than $1$ (corresponding to a $60$ degree angle). In other words, $\psi$ encourages vertices linked by a transition edge to have embeddings with an euclidean distance of $1$, but does not push for a larger difference.

Note that a loss of $0$ can be achieved by any embeddings which are exactly constant within ground truth segments and have a difference of at least $1$ between adjacent segments. The four-color theorem \citep{gonthier2008formal} tells us that this is always possible as long as the dimension of the embeddings is at least $3$. However, because embeddings are computed by $\xi$, which operates on vertex features, there needs to be a recognizable pattern at the border between segments in order for our method to detect a transition.

\textbf{Cross-partition Weighting.}
Note that without an appropriate edge weighting scheme, this loss will only encourage high accuracy in recovering transition edges. However, the influence of each edge can be vastly different in terms of  undersegmentation error $\cL$. Indeed, a single missed edge can result in the erroneous fusion of large adjacent segments. In order for $\ell$ to better represent $\cL$, we choose the edge weights $\mu^{(e)} \in \bbR_+^{\Eintra(\cP)}$ to reflect this influence.

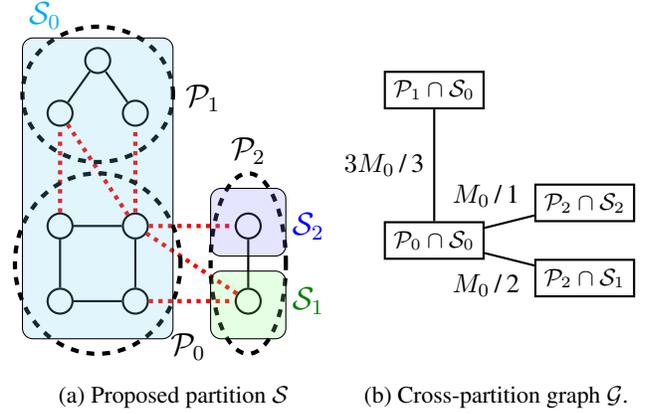
\begin{figure}
\setlength\tabcolsep{0pt}
    %auto-ignore
\begin{tabular}{cc}
 \begin{subfigure}[b]{0.25\textwidth}
 \begin{tikzpicture}
    \tikzstyle{vertex}=[circle, draw = black, thick, minimum size = 3mm]
    \tikzstyle{edge}=[thick, draw = black]
    \tikzstyle{trans}=[ultra thick, draw = red, dotted]

        \node at (0,0) [vertex,fill = white] (v0){};
        \node at (1,0) [vertex,fill = white] (v1){};
        \node at (1,1) [vertex,fill = white] (v2){};
        \node at (0,1) [vertex,fill = white] (v3){};
        
        \node at (0,2.5) [vertex,fill = white] (v4){};
        \node at (1,2.5) [vertex,fill = white] (v5){};
        \node at (0.5,3.2) [vertex,fill = white] (v6){};
        
        \node at (2.5,0) [vertex,fill = white] (v7){};
        \node at (2.5,1) [vertex,fill = white] (v8){};
        
        \draw [edge] (v0) -- (v1) -- (v2) -- (v3) -- (v0);
        \draw [edge] (v8) -- (v7);
        \draw [edge] (v4) -- (v6) -- (v5);
        \draw [trans] (v1) -- (v7) -- (v2) -- (v8);
        \draw [trans] (v3) -- (v4) -- (v2) -- (v5);
        
        \draw [dashed, ultra thick] (0.5,0.5) ellipse (1.1cm and 1.2cm);
        \draw [dashed, ultra thick] (0.5,2.75) ellipse (1cm and .9cm);
        \draw [dashed, ultra thick] (2.5,0.5) ellipse (0.5cm and 1.2cm);
        
        \node at (1.7,-0.6) [draw = none] (l0){\large $\cP_0$};
        \node at (1.9,2.7) [draw = none] (l1){\large $\cP_1$};
        \node at (2.5,2) [draw = none] (l2){\large  $\cP_2$};
        
        \draw [rounded corners, fill = cyan!50!white, fill opacity = 0.2] (-0.5,-0.5) rectangle (1.5,3.5);
        
         \draw [rounded corners, fill = green!50!white, fill opacity = 0.2] (2.0,-0.5) rectangle (3.0,0.4);
         
         \draw [rounded corners, fill = blue!50!white, fill opacity = 0.2] (2.0,0.6) rectangle (3.0,1.5);
         
        \node at (-0.2,3.8) [draw = none, text = cyan] (l0){\large $\cS_0$};
        \node at (3.3,0) [draw = none, text = green!50!black] (l1){\large $\cS_1$};
        \node at (3.3,1) [draw = none, text = blue] (l2){\large  $\cS_2$};
         
    \end{tikzpicture}
    \caption{Proposed partition $\cS$}
     \label{fig:cS}
    \end{subfigure}
    &
    \begin{subfigure}[b]{0.25\textwidth}
      \begin{tikzpicture}
    \tikzstyle{vertex}=[draw = black, thick]
    \tikzstyle{edge}=[thick, draw = black]
    
    \node at (1.5,3.0) [vertex,fill = white] (c0){\small $\cP_1 \cap \cS_0$};
    \node at (1.5,1.0) [vertex,fill = white] (c1){\small$\cP_0 \cap \cS_0$};
    \node at (3.5,1.5) [vertex,fill = white] (c2){\small$\cP_2 \cap \cS_2$};
    \node at (3.5,0.5) [vertex,fill = white] (c3){\small$\cP_2 \cap \cS_1$};
    
     \draw [edge] (c0) -- (c1) node [pos=0.5, fill=none, left] {$3M_0$\,/\,3};
     \draw [edge] (c1) -- (c2) node [pos=0.9, fill=none, above left] {$M_0$\,/\,1};
     \draw [edge] (c1) -- (c3) node [pos=0.9, fill=none, below left] {$M_0$\,/\,2};
    
    \draw [ultra thick, draw = none] (0.5,0.5) ellipse (0.1cm and 1.2cm);
        \draw [ultra thick, draw = none] (0.5,2.75) ellipse (0.1cm and .9cm);
        \draw [ultra thick, draw = none] (2.5,0.5) ellipse (0.1cm and 1.2cm);
        
    \end{tikzpicture}
        \caption{Cross-partition graph $\cG$.}
        \label{fig:cg}
   \end{subfigure}
    \end{tabular}
    \caption{Illustration of the proposed superedge weighting scheme. In \Subref{fig:cS}, we represent an erroneous proposed partition $\cS$ with $3$ segment. In \Subref{fig:cg} we represent the cross partition graph $\cG$ with the edge weights $M_{U,V}/\Abs{U\times V \cap \Etra}$.}
    \label{fig:xseg}
\end{figure}

To this end, we introduce the \emph{cross-partition graph} $\cG=(\cC,\cE)$, represented in \figref{fig:xseg} and defined as the adjacency graph of the cross-partition between  $\cP$ and $\GMPP{e}$ considering only transition edges $\Etra(\cP)$:
\begin{align}\nonumber
\cC &= \Cur{P \cap S\mid P \in \cP, S \in \GMPP{e}\;\text{and}\; P \cap S \neq \varnothing}\\\nonumber
%\cE &= \Cur{\Cur{(u,v) \in \Pa{U \times V} \cap \Etra(\cP)} \mid U,V \in \cC}.
\cE &= \Cur{(U,V) \in \cC^2 \mid U \times V \cap \Etra \neq \varnothing}.
\end{align}
We associate the following weight $M_{U,V}$ to each edge $(U,V)$ of $\cE$ and $\mu_{u,v}$ to each transition edge:
\begin{eqnarray}\nonumber
M_{U,V}^{(e)} \!\!&=&\!\! 
M_0 \min\Pa{\Abs{U}, \Abs{V}} \}\; \text{for}\; (U,V) \in \cE \\\nonumber
\mu_{u,v}^{(e)} \!\!&=&\!\! \frac{M_{U,V}^{(e)}}{\mid U \times V \cap \Etra \mid} \;\text{for}\;(u,v) \in U \times V \cap \Etra.
\end{eqnarray}
with $\mu$ a parameter of the model.
By definition of $\cE$, two segments $U$ and $V$ of $\cC$ linked by a  superedge $(U,V)\in \cE$ are in two different ground truth segments. The undersegmentation error caused by the erroneous fusion of these two components is proportional to $\mini(\Abs{U},\Abs{V})$. This error, spread evenly over the edges constituting the transition superedge, determines the edge weights.

This weighting scheme differs from the SEAL strategy \citep{liu2018learning}. Indeed, in the latter, the edge weights are shared by all transition edges of a given segment. This favors long interfaces in the loss too strongly, and inadequately handles large segments with multiple interfaces. Finally, SEAL sets the weights to $1$ as soon as a border is retrieved, which makes the loss rather unstable and hard to optimize.

Setting $M_0=\Abs{E}/\Abs{V}$ gives the same importance to the classification of transition and non-transition edges. Indeed, assuming that most edges are non-transition, we have the sum of non-transition edge weights close to $\Abs{E}$, while the sum of transition edge weights is in the order of magnitude of $\Abs{V}$. In an oversegmentation setting, $M_0$ must be set higher to prioritize recovering object borders.
%.........................................................................
\begin{figure*}[t]
   \pgfplotsset{
% override style for non-boxed plots
    % which is the case for both sub-plots
    every non boxed x axis/.style={} 
}

\setlength{\belowcaptionskip}{-5pt}
\setlength{\abovecaptionskip}{-5pt}
\pgfplotstableread{./data/ptn_s3dis.txt}{\ptnstanford}
\pgfplotstableread{./data/cvpr_s3dis.txt}{\cvprstanford}
\pgfplotstableread{./data/isprs_s3dis.txt}{\isprsstanford}
\pgfplotstableread{./data/geof_s3dis.txt}{\geofstanford}
\pgfplotstableread{./data/slic_s3dis.txt}{\slicstanford}
\pgfplotstableread{./data/ptn_seal_s3dis.txt}{\sealstanford}

\pgfplotstableread{./data/ptn_vkitticolor.txt}{\ptnvkitti}
\pgfplotstableread{./data/ptn_vkitticolor2.txt}{\ptnvkittid}
\pgfplotstableread{./data/cvpr_vkitti.txt}{\cvprkitti}
\pgfplotstableread{./data/isprs_vkitti.txt}{\isprskitti}
\pgfplotstableread{./data/geof_vkitti.txt}{\geofkitti}
\pgfplotstableread{./data/slic_vkitti.txt}{\slicvkitti}
\pgfplotstableread{./data/ptn_seal_vkitti.txt}{\sealvkitti}

\pgfplotstableread{./data/ptn_vkittinocolor.txt}{\ptnvkittinocolor}
\pgfplotstableread{./data/isprs_vkitti.txt}{\isprskittinocolor}
\pgfplotstableread{./data/geof_vkittinocolor.txt}{\geofkittinocolor}

 \tikzstyle{ptn}=[black, ultra thick]
 \tikzstyle{geof}=[green!70!black, ultra thick, dashed]
  \tikzstyle{slic}=[cyan, ultra thick]
 \tikzstyle{cvpr}=[red, ultra thick]
 \tikzstyle{isprs}=[blue, ultra thick, dotted]
  \tikzstyle{seal}=[pink, ultra thick, dotted]
 \setlength\tabcolsep{0pt}
\begin{center}
\begin{tabular}{ccc}
%\begin{multirow}{2}{*}{
 
  %}
  %&
  \setcounter{subfigure}{0}
  \begin{subfigure}[b]{.32\textwidth} %\centering
  \resizebox{1\textwidth}{!}{
    \begin{tikzpicture}
      \begin{axis}
        [xmin = 300, xmax = 1000, xlabel={\# segments}, ylabel={$OOA$} , ymin = 92, ymax = 98, y label style={at={(+0.05,0.5)}}]
        \addplot [ptn] table [x = {N}, y = {ASA}] {\ptnstanford};
        \addplot [slic] table [x = {N}, y = {ASA}] {\slicstanford};
        \addplot [seal] table [x = {N}, y = {ASA}] {\sealstanford};
        \addplot [geof] table [x = {N}, y = {ASA}] {\geofstanford};
        \addplot [cvpr] table [x = {N}, y = {ASA}] {\cvprstanford};
        \addplot [isprs] table [x = {N}, y = {ASA}] {\isprsstanford};
      \end{axis}
    \end{tikzpicture}
    }
    \caption{OOA for S3DIS}
    \label{fig:OA:S3DIS}
  \end{subfigure}&
   
\begin{subfigure}[b]{0.31\textwidth} %\centering
\resizebox{1\textwidth}{!}{
    \begin{tikzpicture}
      \begin{axis}
        [xmin = 300, xmax = 1000, xlabel={\# segments}, ylabel={$BR$}, ymin = 45, ymax = 85, y label style={at={(+0.05,0.5)}}]
        \addplot [ptn] table [x = {N}, y = {BR}] {\ptnstanford};
        \addplot [slic] table [x = {N}, y = {BR}] {\slicstanford};
         \addplot [seal] table [x = {N}, y = {BR}] {\sealstanford};
       \addplot [geof] table [x = {N}, y = {BR}] {\geofstanford};
           \addplot [cvpr] table [x = {N}, y = {BR}] {\cvprstanford};
        \addplot [isprs] table [x = {N}, y = {BR}] {\isprsstanford};
      \end{axis}
    \end{tikzpicture}
    }
       \caption{BR for S3DIS}
       \label{fig:BR:S3DIS}
  \end{subfigure} 
  &
\begin{subfigure}[b]{.32\textwidth} %\centering
\resizebox{1\textwidth}{!}{
    \begin{tikzpicture}
      \begin{axis}
        [xmin = 300, xmax = 1000, xlabel={\# segments}, ylabel={$BP$} , ymin = 10, ymax = 30, y label style={at={(+0.05,0.5)}}]

        \addplot [ptn] table [x = {N}, y = {BP}] {\ptnstanford};
        \addplot [slic] table [x = {N}, y = {BP}] {\slicstanford};
         \addplot [seal] table [x = {N}, y = {BP}] {\sealstanford};
        \addplot [geof] table [x = {N}, y = {BP}] {\geofstanford};
        \addplot [cvpr] table [x = {N}, y = {BP}] {\cvprstanford};
        \addplot [isprs] table [x = {N}, y = {BP}] {\isprsstanford};
        \addlegendentry{\large SSP (Ours)}
        \addlegendentry{\large SSP-Cluster (Ours)}
         \addlegendentry{\large SSP-SEAL(Ours)}
        \addlegendentry{\large Geom-Graph}
        \addlegendentry{\large VCCS}
        \addlegendentry{\large Lin \etal}
      \end{axis}
    \end{tikzpicture}
     }
      \caption{BP for S3DIS}
      \label{fig:BP:S3DIS}
  \end{subfigure}
  \\
  
 %  \begin{subfigure}[b]{.25\textwidth%} %\centering
 % \resizebox{1\textwidth}{!}{
   % \begin{tikzpicture}
     % \begin{axis}
       % [xmin = 300, xmax = 1000, %xlabel={\# segments}, %ylabel={BR} , ymin = 50, ymax %= 95, y label %style={at={(+0.05,0.5)}}]
       % \addplot [ptn] table [x = %{N}, y = {BR}] %{\ptnvkittinocolor};
        %\addplot [cvpr] table [x = {N}, y = {ASA}] {\cvprstanford};
       % \addplot [isprs] table [x = %{N}, y = {BR}] %{\isprskittinocolor};
       % \addplot [geof, mark=x] table %[x = {N}, y = {BR}] %{\geofkittinocolor};
  %    \end{axis}
  %  \end{tikzpicture}
  %  }
  %  \caption{BR for vKITTI no color}
  %  \label{fig:OA:VKITTI}
  %\end{subfigure}
  %&
  \if 1 0
 \begin{subfigure}[b]{.32\textwidth} %\centering
 \resizebox{1\textwidth}{!}{
\begin{tikzpicture}

\begin{groupplot}[
    group style={
        group name=my fancy plots,
       group size=1 by 2,
       xticklabels at=edge bottom,
       vertical sep=0pt
    },
    width=5.5cm,
    xmin=300, xmax=1000
]
\nextgroupplot[ymin=95,ymax=100,
               ytick={96,98,99,100},
               axis x line=top, 
               axis y discontinuity=parallel,
               height=3.5cm]
 %\addplot [cvpr] table [x = {N}, y = {ASA}] {\cvprkitti};
\addplot [isprs] table [x = {N}, y = {ASA}] {\isprskitti};        
\addplot [black] {99+0.001*x};

\nextgroupplot[ymin=80,ymax=87,
               ytick={80,85},
               axis x line=bottom,
               height=3.0cm]
\addplot [cvpr] table [x = {N}, y = {ASA}] {\cvprkitti};
\addplot [isprs] {99 + x * 0.001};
%\addplot [isprs] table [x = {N}, y = {ASA}] {\isprskitti};  
   %width=3.5cm,
  % xmin = 300, xmax = 1000,
   %xlabel={\# segments}, ylabel={$OOA$},
  % y label style={at={(+0.05,0.5)}}
  %  ]

%\nextgroupplot[ymin=95,ymax=100,
 %             ytick={95,100},
  %            axis x line=top, 
  %             axis y discontinuity=parallel,
  %             height=5cm]
  %             
  %\addplot [cvpr] table [x = {N}, y = {ASA}] {\cvprkitti}; %\addplot [isprs] table [x = {N}, y = {ASA}] {\isprskitti};
  % 
   %
  % \nextgroupplot[ymin=80,ymax=90,
   %            ytick={80,85},
   %            axis x line=bottom,
    %           height=2cm]
   %%\addplot [cvpr] table [x = {N}, y = {ASA}] {\cvprkitti}; %\addplot [isprs] table [x = {N}, y = {ASA}] %{\isprskitti}; 
\end{groupplot}
\end{tikzpicture}
}
\end{subfigure}
\fi

\begin{subfigure}[b]{0.31\textwidth} %\centering
\resizebox{1\textwidth}{!}{
    \begin{tikzpicture}
      \begin{axis}
        [xmin = 300, xmax = 1000, xlabel={\# segments}, ylabel={$OOA$}, ymin = 93, ymax = 99, y label style={at={(+0.05,0.5)}}]
	
       \addplot [ptn] table [x = {N}, y = {ASA}] {\ptnvkitti};
        \addplot [slic] table [x = {N}, y = {ASA}] {\slicvkitti};
        \addplot [geof, mark=x] table [x = {N}, y = {ASA}] {\geofkitti};
        \addplot [cvpr] table [x = {N}, y = {ASA}] {\cvprkitti};
        \addplot [isprs] table [x = {N}, y = {ASA}] {\isprskitti};
         \addplot [seal] table [x = {N}, y = {ASA}] {\sealvkitti};
      \end{axis}
    \end{tikzpicture}
    }
       \caption{OOA for vKITTI}
       \label{fig:OA:VKITTIRGB}
  \end{subfigure} 
  &
\begin{subfigure}[b]{0.31\textwidth} %\centering
\resizebox{1\textwidth}{!}{
    \begin{tikzpicture}
      \begin{axis}
        [xmin = 300, xmax = 1000, xlabel={\# segments}, ylabel={$BR$}, ymin = 60, ymax = 96, y label style={at={(+0.05,0.5)}}]
	 %\addplot [ptn] table [x = {N}, y = {BR}] {\ptnvkittid};
       \addplot [ptn] table [x = {N}, y = {BR}] {\ptnvkitti};
         \addplot [slic] table [x = {N}, y = {BR}] {\slicvkitti};
       \addplot [geof, mark=x] table [x = {N}, y = {BR}] {\geofkitti};
           \addplot [cvpr] table [x = {N}, y = {BR}] {\cvprkitti};
        \addplot [isprs] table [x = {N}, y = {BR}] {\isprskitti};
        \addplot [seal] table [x = {N}, y = {BR}] {\sealvkitti};
      \end{axis}
    \end{tikzpicture}
    }
       \caption{BR for vKITTI}
       \label{fig:BR:VKITTIRGB}
  \end{subfigure} 
  &
\begin{subfigure}[b]{.31\textwidth} %\centering
\resizebox{1\textwidth}{!}{
    \begin{tikzpicture}
      \begin{axis}
        [xmin = 300, xmax = 1000, xlabel={\# segments}, ylabel={$BP$} , ymin = 15, ymax = 65, y label style={at={(+0.05,0.5)}}]

        \addplot [ptn] table [x = {N}, y = {BP}] {\ptnvkitti};
         %\addplot [ptn] table [x = {N}, y = {BP}] {\ptnvkittid};
        \addplot [geof, mark=x] table [x = {N}, y = {BP}] {\geofkitti};
          \addplot [slic] table [x = {N}, y = {BP}] {\slicvkitti};
        \addplot [cvpr] table [x = {N}, y = {BP}] {\cvprkitti};
        \addplot [isprs] table [x = {N}, y = {BP}] {\isprskitti};
          \addplot [seal] table [x = {N}, y = {BP}] {\sealvkitti};
      \end{axis}
    \end{tikzpicture}
     }
     \caption{BP for vKITTI}
     \label{fig:BP:VKITTIRGB}
  \end{subfigure}
 
\end{tabular}
\end{center}
 \caption{Performance of the different algorithms on the $6$-fold S3DIS dataset (\subref{fig:OA:S3DIS}, \subref{fig:BR:S3DIS}, \subref{fig:BP:S3DIS}), %the vKITTI dataset without color (\subref{fig:OA:VKITTI}) 
 and the $6$-fold vKITTI3D dataset (\subref{fig:OA:VKITTIRGB},  \subref{fig:BR:VKITTIRGB}, \subref{fig:BP:VKITTIRGB}).}
 \label{fig:spseg}
\end{figure*}
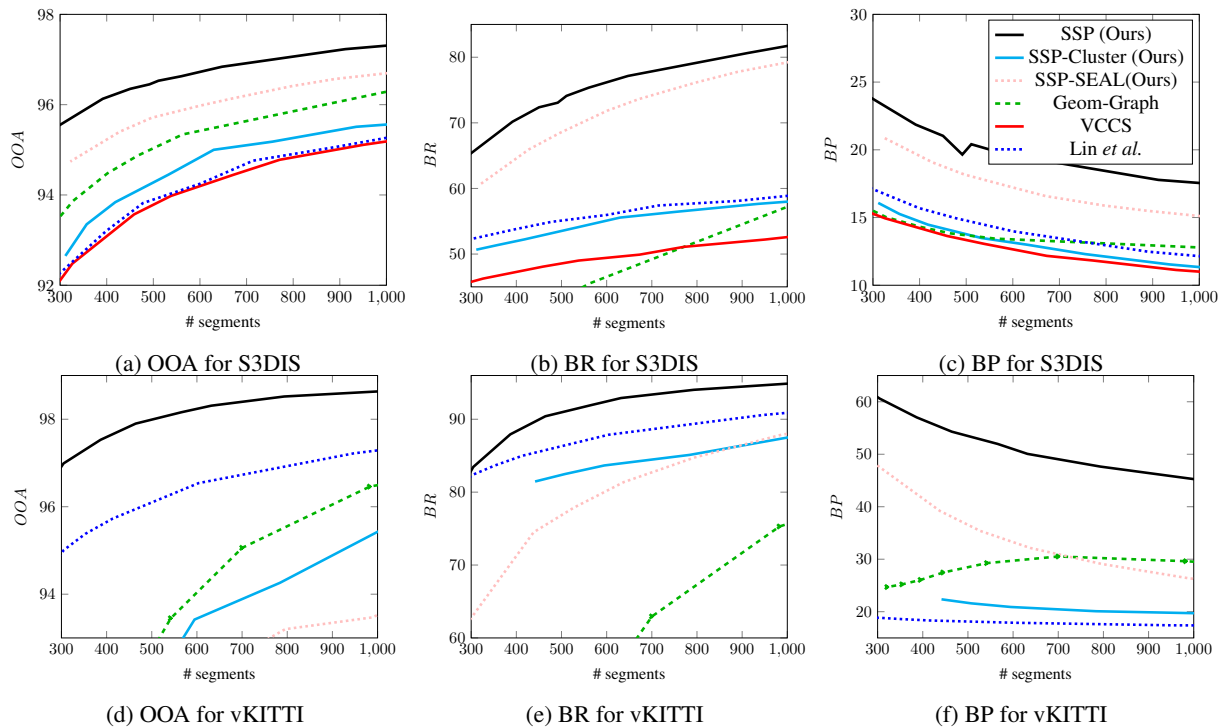
\vspace{-3mm}
\section{Numerical Experiments}
\vspace{-2mm}
\label{sec:xp}
We present numerical illustrations of our approach for 3D point cloud oversegmentation. To this end we consider two different datasets: S3DIS, composed of  dense indoor scans \citep{Armeni16_s3dis}, and vKITTI3D, a virtual dataset of sparse outdoor point clouds \citep{EngelmannKHL17_vkitty, Gaidon:Virtual:CVPR2016}. The first one has both object and semantic label annotations. For the second one, we define the ground truth partition $\cP$ as the connected components of the semantic labels. Both datasets are composed of $6$ independent parts, which allows us to perform $6$-fold cross-validation.
Note that for considerations of simplicity and efficiency, we relax the optimization domain of \eqref{eq:mgp} from $\bbS_m$ to $\bbR^m$, while still learning spherical embeddings. While this can lead to suboptimal partitions, the segmentation retrieved are still relevant. We set the dimensions of the embeddings to $4$, and $M_0$ to $5\Abs{E}/\Abs{V}$, hence favoring oversegmentations.
The embedding function is a small PointNet-like network \citep{qi2017pointnet} operating on the $20$ nearest neighbors of each point. More details in the appendix. %We remove the input transform and bypass the following arguments directly to the maxpooled map: $z$ coordinate of the point, scaling factor when normalizing the neighborhood point clouds, and the rotation matrix outputted by the spatial transform.

In \figref{fig:spseg}, we report the performance of our algorithm according to three segmentation metrics: OOA, BR, and BP. OOA denotes the Oracle Overall Accuracy, \ie the OA of the oracle classification algorithm associating the majority label to each segment of the proposed partition $\cS$. Note that the OOA is a higher bound on the pointwise overall accuracy of any classification algorithm operating on the segments. The OOA is also closely linked to the undersegmentation error, but adds a semantic component. BP (resp. BR) denotes the precision (resp. recall) of the predicted transition edges $\Eintra(\cS)$ compared to $\Eintra(\cP)$ with a tolerance of one edge.
We denote our method by \textbf{SSP} for \emph{Supervized SuperPoints} and compare our approach to the following methods:
\vspace{-4mm}
\begin{itemize}[leftmargin=2mm]
\setlength\itemsep{0mm}
\setlength{\itemindent}{0mm}
\item[]\textbf{SSP-cluster} is our adaptation of the soft partition approach of \citet{JampaniSLYK18} to the 3D setting.
\item[]\textbf{SSP-SEAL} uses the same framework as \textbf{SSP}, but with the cross-partition weights replaced by the SEAL weighting strategy \citep{liu2018learning}. Note that this is \emph{not} equivalent to the framework of \citet{liu2018learning}, as they use a different loss and clustering algorithm.
\item[]\textbf{Geom-graph} is the graph-based method introduced by \citet{guinard2017weakly} solving \eqref{eq:mgp} on handcrafted features \citep{demantke2011dimensionality} instead of learned ones.
\item[]\textbf{VCCS} is the octree-structured cluster-based method introduced by \citet{PaponASW13}.
\item[]\textbf{Lin \etal} is the adaptive resolution graph-based method introduced by \citet{LIN201839}.
\end{itemize}
\vspace{-3mm}
We observe that for the large S3DIS dataset ($600$ Mpoints), supervized methods provide considerably better results. In particular, our method \textbf{SSP} obtains better accuracy with $300$ segments than the state-of-the-art method of \textbf{Lin \etal} with $1500$ segments. The advantages for border recall and precision are even more significant. For the smaller vKITTI3D dataset ($15$ Mpoints), \textbf{Lin \etal}\!\! obtain better results than all supervized methods except our approach. Illustration of the results as well as more details on the models and metrics are given in the appendix.
\vspace{-3mm}
\section*{Conclusion}
\vspace{-2mm}
We presented a framework for learning to segment graph-structured data with neural networks. Our new loss is fully backpropagable and indirectly takes the undersegmentation error into account. We assess its efficiency on two large-scale point cloud oversegmentation benchmarks. We demonstrate a significant improvement over unsupervized methods of the state-of-the-art and our own implementations of other supervized methods. All codes will be released at the following URL: \url{github.com/loicland/superpoint_graph}. Future works includes applying our method to other graph-structured data types such as images or relationship graphs and solving the GMPP with the spherical domain constraint.
%%%%%%%%%%%%%%%%%%%%%%%%%%%%%%%%%%%%%%%%%%%%%%%%%%%%%%%%%%%%%%%%%%%%%%%%%%%%%%%
%%%%%%%%%%%%%%%%%%%%%%%%%%%%%%%%%%%%%%%%%%%%%%%%%%%%%%%%%%%%%%%%%%%%%%%%%%%%%%%
%\pagebreak
%\clearpage
\balance
{\small
\bibliographystyle{icml2019}
%\bibliography{mybib}

}
\pagebreak
\clearpage

%.....................................................................
\appendix
\section*{SUPPLEMENTARY MATERIALS}
\setcounter{section}{0}

\setlength\tabcolsep{0pt}
\begin{figure*}[!ht]\centering
\begin{tabular}{cc}
\begin{subfigure}[b]{0.4\textwidth}
\includegraphics[width=1\textwidth]{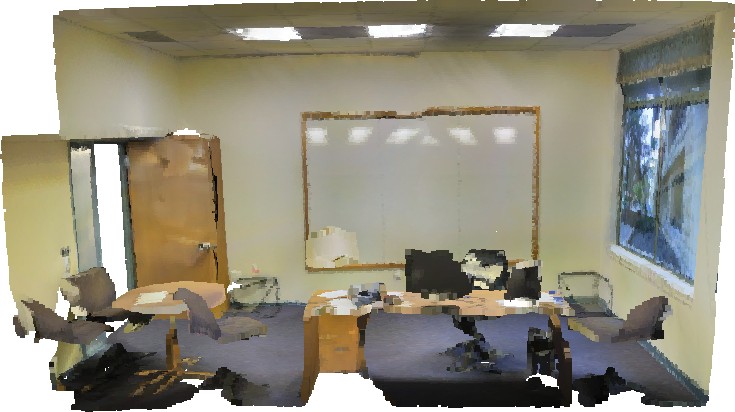}
\caption{Input Point Cloud}
\label{fig:introfig_rgb}
\end{subfigure}
&
\begin{subfigure}[b]{0.4\textwidth}
\includegraphics[width=1\textwidth]{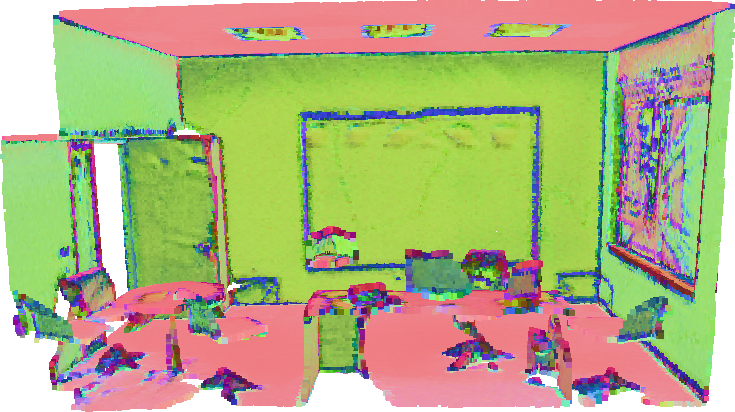}
 \caption{Learned Embedding}
\label{fig:introfig_emb}
\end{subfigure}
\\
\begin{subfigure}[b]{0.4\textwidth}
\includegraphics[width=1\textwidth]{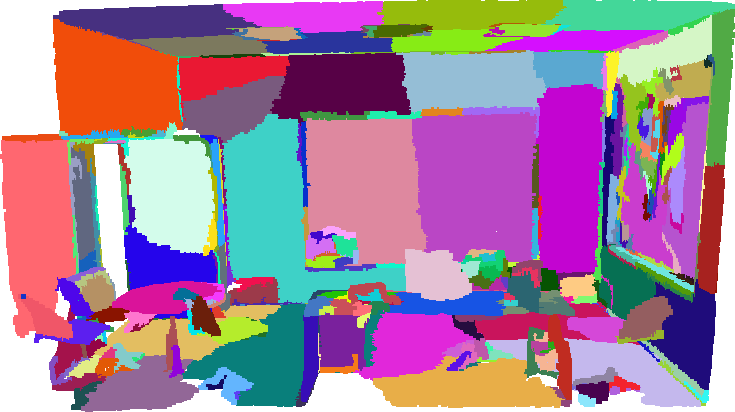}
\caption{Oversegmentation}
\label{fig:introfig_seg}
\end{subfigure}
&
\begin{subfigure}[b]{0.4\textwidth}
\includegraphics[width=1\textwidth]{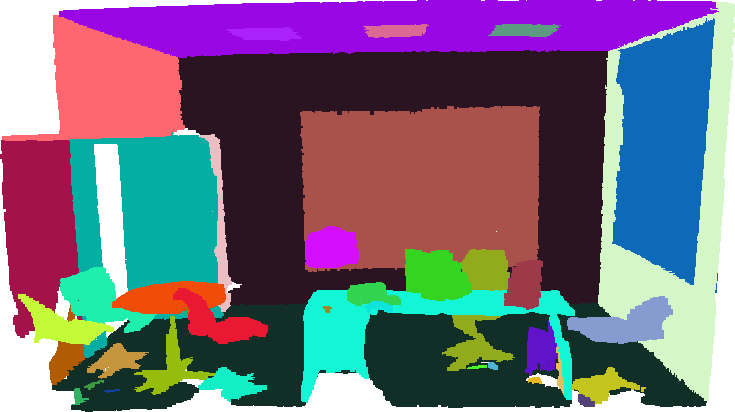}
\caption{True Objects}
\label{fig:introfig_gt}
\end{subfigure}
\end{tabular}
\caption{Illustration of our framework on a hard-to-segment scene with a white board on a white wall: a colored point cloud is given as input \Subref{fig:introfig_rgb}, an embedding is computed for each point \Subref{fig:introfig_emb}, which allows a clustering technique to compute an oversegmentation \Subref{fig:introfig_seg}, which closely follows the ground truth \Subref{fig:introfig_gt}. The embeddings are projected into a 3-dimensional space to allow color visualization.
}
\label{fig:illu_all}
\end{figure*}

%......................................................
\section{3D Point Embedding }
%.....................................................................
\label{sec:emb}
In this section, we describe the embedding function $\xi$ used in the 3D point cloud oversegmentation application.
This function associate to each point a spherical $m$-dimensional embedding $e_i$ characterizing its point-features (position, color, etc.) and the geometry and radiometry of its local neighborhood. 
To this end, we introduce the Local Point Embedder ($\LCE$), a lightweight network inspired by PointNet \cite{qi2017pointnet}. However, unlike PointNet, $\LCE$ does not try to extract information from the whole input point cloud, but rather encodes each point based on purely local information. Here, we describe the different units of our network.\\
\textbf{Spatial Transform:} This unit takes the positions of a target point $p_i$ and its local $k$-neighborhood $P_i$. It normalizes the neighbors' coordinates around $p_i$, and such that the standard deviation of the point's position is equal to $1$ \eqref{eq:norm}. Then, this neighborhood is rotated around the $z$ axis with a $2\times2$ rotation matrix computed by small PointNet network $\PTN$ \eqref{eq:rot}. As advocated by \cite{jaderberg2015spatial}, these steps aim to standardize the position of the neighborhood clouds of each point. This helps the next network to learn position distribution. 
Along the normalized neighborhood position $\Pit$, this unit also outputs geometric point-features $\pit$ describing the elevation $\upp{p}{z}_i$, the neighborhood radius, as well as its original orientation (through the $4$ values of the rotation matrix: $[\Omega_{x,x},\Omega_{x,y},\Omega_{y,x},\Omega_{y,y}]$) \eqref{eq:pout}. By keeping track of the normalization operations, the embedding can stay covariant with the original neighborhood's radius, height, and original orientation, even though the points' positions have been normalized and rotated.
\begin{eqnarray}
 \rad  &=&  \std{P_i} \label{eq:diam}
 \\\label{eq:st}
 \Omega &=& \PTN(\Pit) 
 \\\label{eq:norm}
 P_i'  &=&  (P_i - p_i) /\rad 
 \\\label{eq:rot}
  \Pit  &=&  \{p \times \Omega \mid p \in P_i'\} 
  \\\label{eq:pout}
 \pit  &=&  [p_i^{(z)}, \rad, \Omega]
\end{eqnarray}
\textbf{Local Point Embedder:} 
The $\LCE$ network computes a normalized embedding from two inputs: a point-feature $x_i$ and a set-feature $X_i$. As in PointNet \cite{qi2017pointnet}, the set-features are first processed independently by a multi-layer perceptron (denoted $\MLP_1$) comprised of a succession of layers in the following order: linear, activation (ReLu \cite{nair2010rectified}), normalization (batch \cite{ioffe2015batch}), and so on. The resulting set-features are then maxpooled into a point-feature, which is concatenated with the input point-feature. The resulting vector is processed through another multi-layer perceptron $\MLP_2$ \eqref{eq:ptn}, and finally normalized on the unit sphere.

The embeddings $e_i$ are computed for each point $i$ of $C$ through a shared $\LCE$ \eqref{eq:emb}. The input set-feature $X_i$ is set as the concatenation of the neighbour's transformed position $\Pit$ and their radiometric information $R_i$, while the input point-feature $x_i$ is composed of the neighborhood geometric point-feature $\pit$ and the radiometry $r_i$ of  point $i$.
\begin{flalign}\label{eq:ltwo}
&\!\!\!\!\!\!\!\!\!\!\Ltwo(\cdot)=\cdot/\Vert \cdot \Vert\\\label{eq:ptn}
&\!\!\!\!\!\!\!\!\!\!\LCE(X_i,x_i)\!=\!\Ltwo\Pa{\MLP_2\Pa{[\max\Pa{\MLP_1(X_i)}, x_i]}}\\\label{eq:emb}
&\!\!\!\!\!\!\!\!\!\!e_i=\LCE([\Pit,R_i],[\pit,r_i])
\end{flalign}

\subsection {Implementation Details}
%....................................................................
\label{sec:details}
We use a modified version of the $\ell_0$-cut pursuit algorithm\footnote{\url{https://github.com/loicland/cut-pursuit}}\cite{landrieu2017cut}, with two main differences:
\begin{itemize}
\item to prevent the creation of many small segments in regions of high contrast, we merge components greedily with respect to the MGPP energy (1), as long as they are smaller than a given threshold;
\item we heuristically improved the forward step (1) from \cite{landrieu2017cut}, such that the regularization strength increases geometrically by a factor (of $0.7$) along the iterations. This helps improve the quality of the lower optima retrieved, and consequently the graph partition.
\end{itemize}

To limit the spatial extent of the segment we concatenate to the points' embeddings their 3D coordinates in (1) multiplied by a parameter $\aspat$, in the manner of \cite{achanta2012slic}. This determines the maximum size that superpoints can reach.

In all our experiments, we set $m$ the dimension of our embeddings to $4$. We choose a light architecture for the $\LCE$, with less than $15,000$ parameters.

%************************

\section{Oversegmentation Metrics}

There are many standard metrics which assess the quality of point cloud oversegmentation.
%with respect to properties \ref{prop:p1}, \ref{prop:p2}, and \ref{prop:p3}.
In particular, the Boundary Recall (BR) and Precision (BP) are used to evaluate the ability of the superpoints to adhere to, and not cross, object boundaries. %(\ref{prop:p2}, \ref{prop:p3}).
In the literature, these measures are defined with respect to \emph{boundary pixels} \cite{PaponASW13} or points \cite{LIN201839}. However, we argue that transition occurs \emph{between} points and not \emph{at} points for point clouds. Consequently, we define $\Etransp$ the set of predicted transition, \ie the subset of edges of $E$ that connect two points of $C$ in two different superpoints. These metrics are often given with respect to a tolerance, \ie the distance at which a predicted transition must take place from an actual object's border for the latter to be considered retrieved. We set this distance to $1$ edge, which leads us to define $\Etranse$ the set of inter-edges expanded to all directly adjacent edges in $E$:
$$\Etranse = \Cur{(i,j) \in E \mid \exists (i,k) \;\text{or}\; (j,k) \in \Etra}.$$
This allows us to define the boundary recall and precision with $1$ edge tolerance for a set of predicted transition $\Etransp$:
$$
BR = \frac{\mid\Etransp \cap \Etranse\mid}{\mid\Etra\mid},\;
BP = \frac{\mid\Etransp \cap \Etranse\mid}{\mid\Etransp\mid}.
$$
%Since the end-goal of our point cloud oversegmentation framework is to provide useful superpoints for semantic segmentation, we define the \emph{Oracle Overall Accuracy} (OOA). 

To assess object purity we define the \emph{Oracle Overall Accuracy} (OOA). This metric characterizes the accuracy of the labeling that associates each superpoint $S$ of a segmentation $\cS$ with its majority ground-truth label. Formally, let $l \in \cK^C$ be the semantic labels of each point within a set of classes $\cK$, we define the OOA of a point cloud segmentation $\cS$ as:
\begin{align}\nonumber
    &\loracle(S)=\text{mode}\Cur{l_i \mid i \in S}\\\nonumber
    &OOA= \frac1{\mid C \mid} \sum_{S \in \cS}\sum_{i \in S} \Bra{l_i=\loracle(S)},
\end{align}
with $[x=y]$ the function equal to $1$ if $x=y$ and $0$ otherwise. Note that the OOA is closely related to the ASA \cite{liu2011entropy}, but consider the majority labels of all points within a superpixel rather than the label of the objects with most overlap. 
%In this sense, it is a tighter upper bound to the achievable accuracy of a superpoint-based semantic classification algorithm using $\cS$. 
This metric is also more fair than the undersegmentation error \cite{levinshtein2009turbopixels} for other methods such as \cite{guinard2017weakly}, or our cluster-based approach, as they do not try to retrieve objects directly, but rather regions of $C$ with homogeneous semantic labeling.

\begin{figure*}
\begin{center}
\begin{tabular}{c}
\begin{subfigure}[b]{1\textwidth}
\begin{tabular}{ccc}
\includegraphics[width=.3\textwidth]{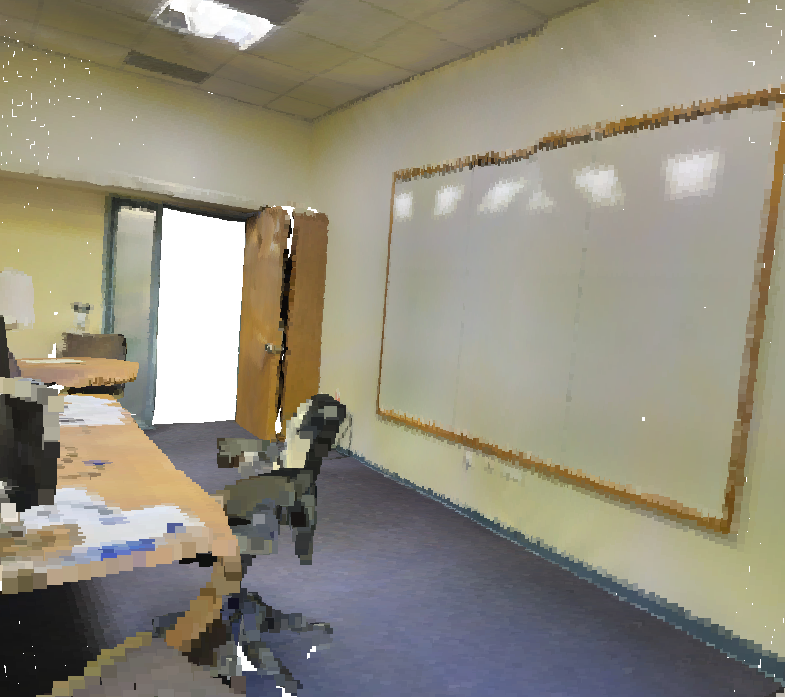}
&
\includegraphics[width=.3\textwidth]{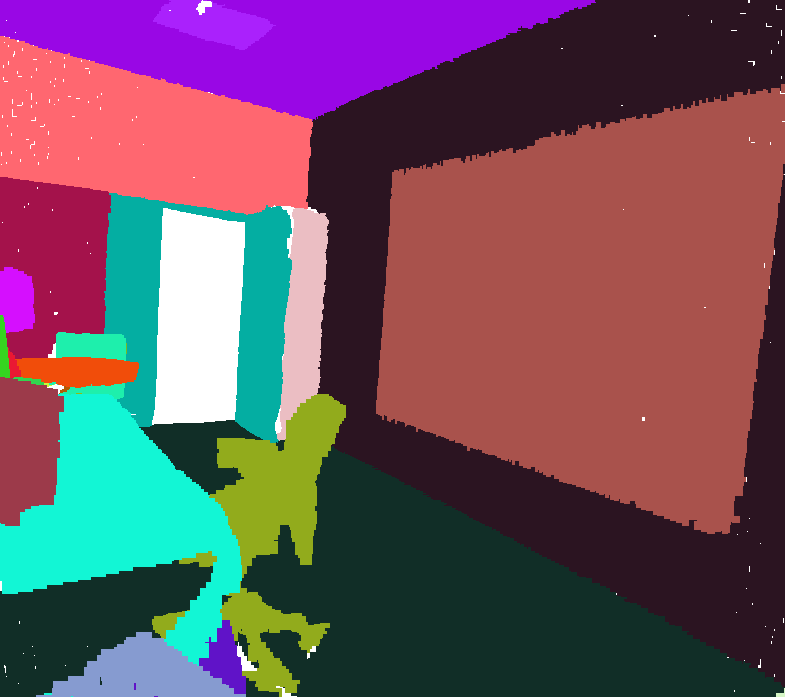}
&
\includegraphics[width=.3\textwidth]{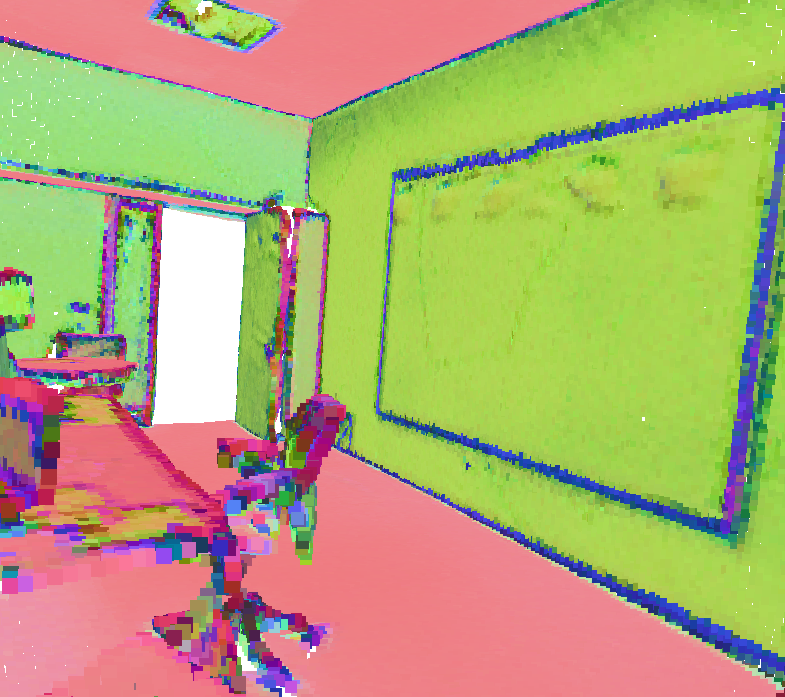}
\\
Input cloud
&
Ground truth objects
&
$\LCE$ embeddings
\\
\includegraphics[width=.3\textwidth]{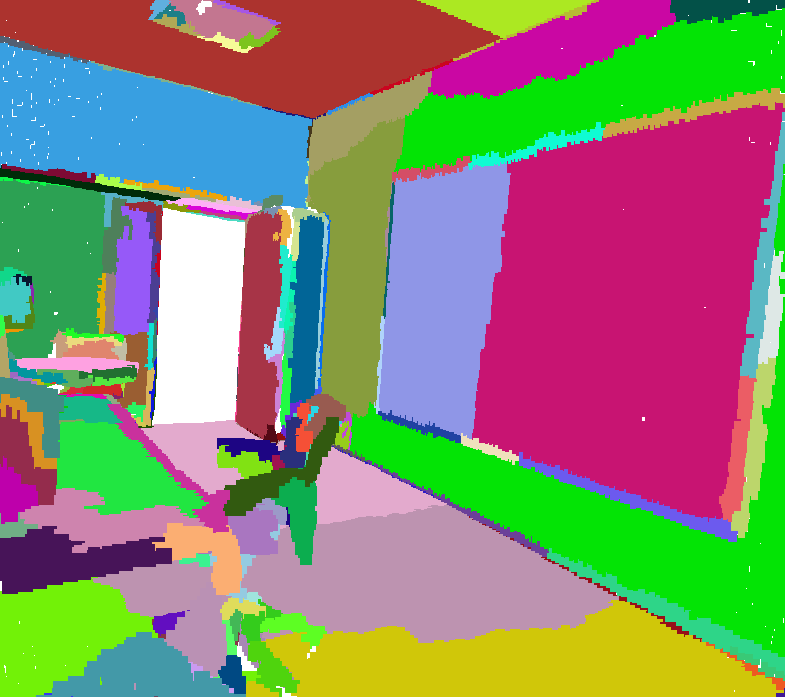}
&
\includegraphics[width=.3\textwidth]{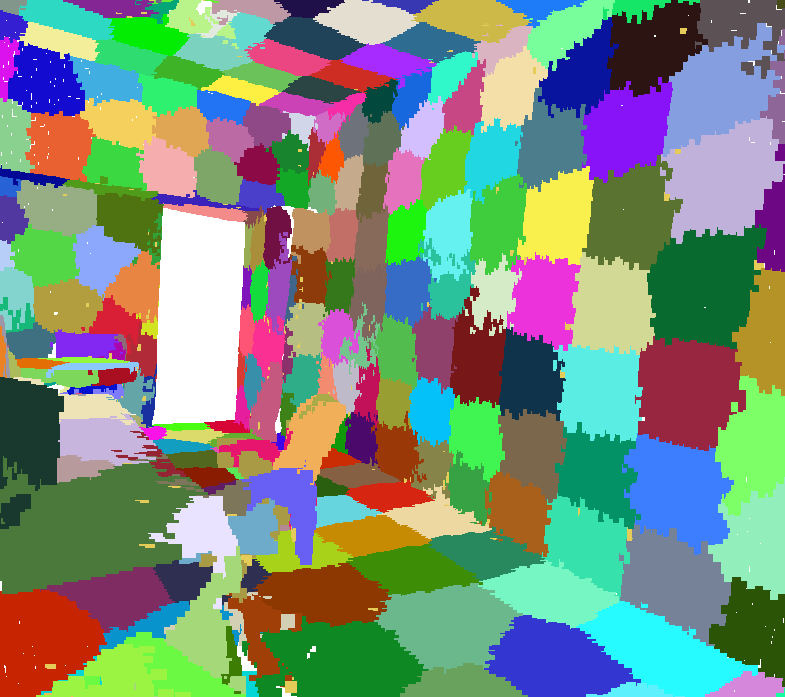}
&
\includegraphics[width=.3\textwidth]{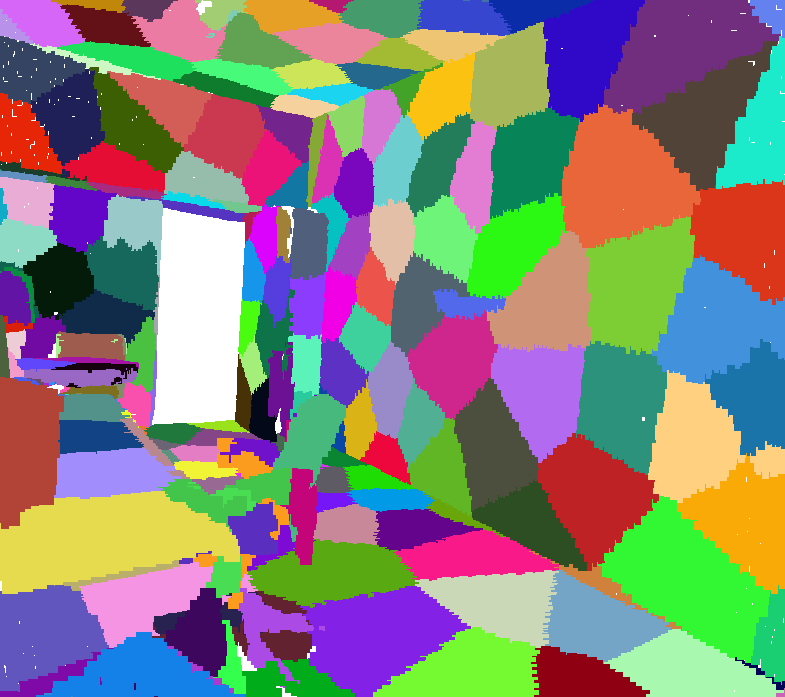}
\\
SSP (ours)
&
VCCS
&
Lin \etal
\end{tabular}
\caption{S3DIS scene with 58 objects. Superpoint count : SSP 442, VCCS 436, Lin 423.}
\vspace{.5cm}
\end{subfigure}
\\
\begin{subfigure}[b]{1\textwidth}
\begin{tabular}{ccc}
\includegraphics[width=.3\textwidth]{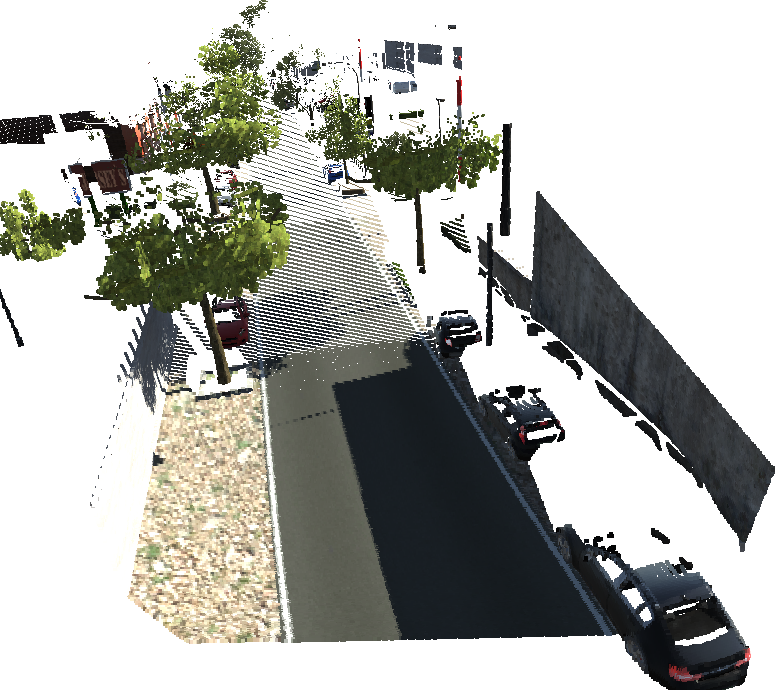}
&
\includegraphics[width=.3\textwidth]{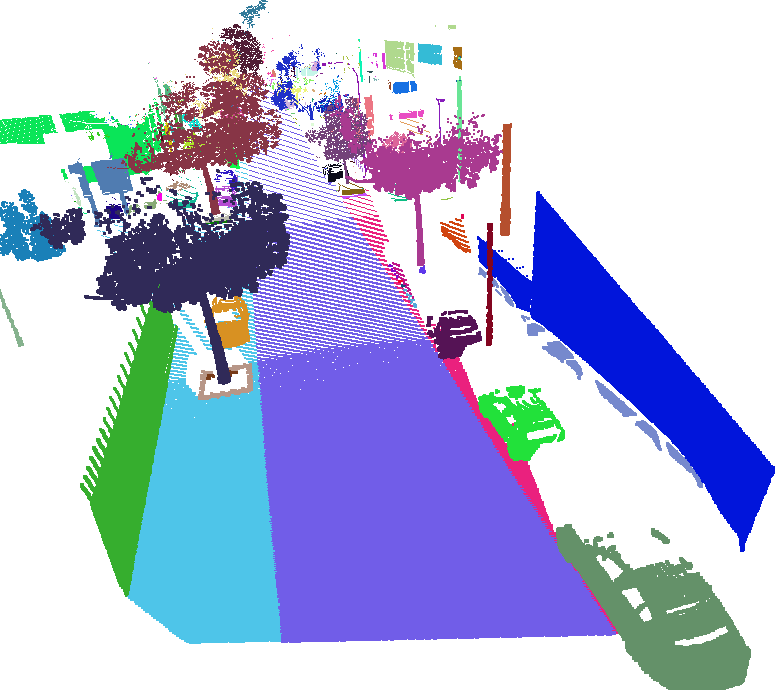}
&
\includegraphics[width=.3\textwidth]{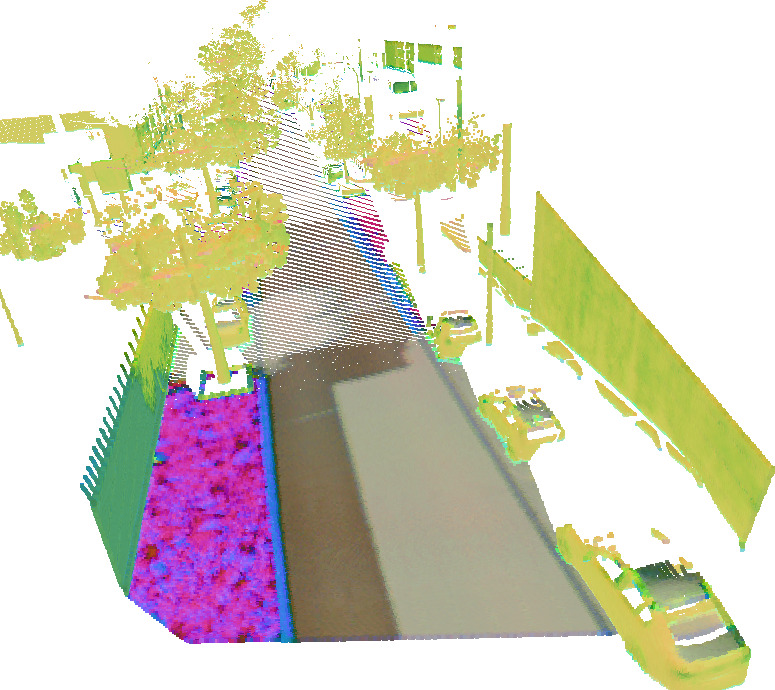}
\\
Input cloud
&
Ground truth objects
&
$\LCE$ embeddings
\\
\includegraphics[width=.3\textwidth]{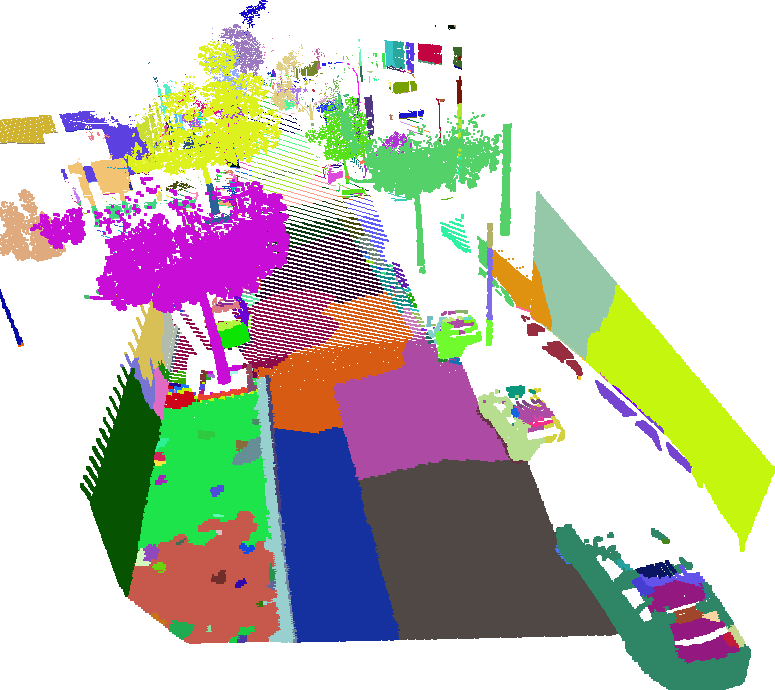}
&
\includegraphics[width=.3\textwidth]{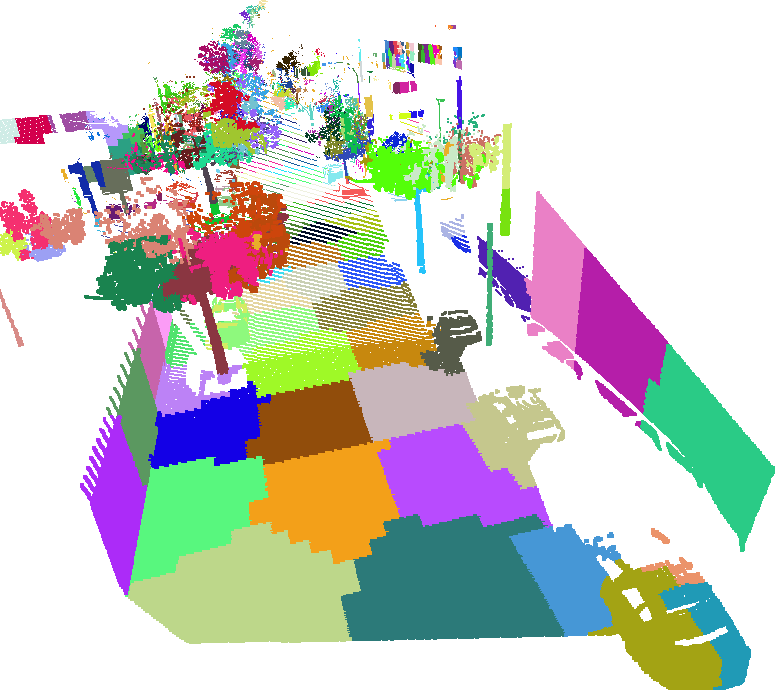}
&
\includegraphics[width=.3\textwidth]{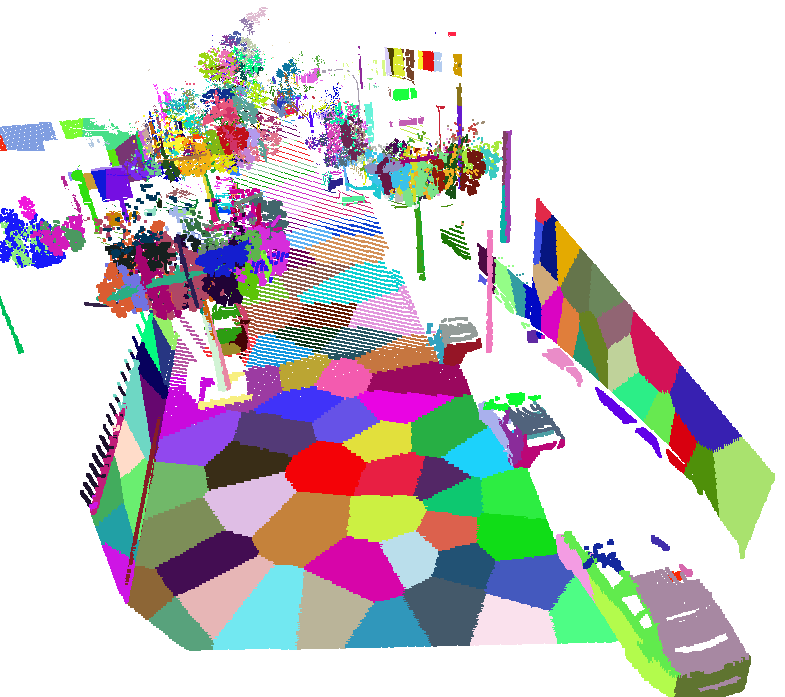}
\\

SSP (ours)
&
VCCS 
&
Lin \etal
\end{tabular}
\caption{vKITTI scene with  $233$ objects. Superpoint count: SSP 420, VCCS 422, Lin 425.}
\end{subfigure}
%\\
%\begin{subfigure}[b]{1\textwidth}
%\begin{tabular}{cccccc}
%\includegraphics[width=.15\textwidth]{./images/ini2.png}
%&
%\includegraphics[width=.15\textwidth]{./images/res2.png}
%&
%\includegraphics[width=.15\textwidth]{./images/tot2.png}
%&
%\includegraphics[width=.15\textwidth]{./images/obj2.png}
%&
%\includegraphics[width=.15\textwidth]{./images/obj2.png}
%&
%\includegraphics[width=.15\textwidth]{./images/obj2.png}
%\end{tabular}
%\caption{vKITTI dataset without color information}
%\end{subfigure}
\end{tabular}
\end{center}
\caption{Illustration of the oversegmentations of our framework, and from competing algorithms.}
\label{fig:illu_res}
\end{figure*}

In Figure~\ref{fig:illu_res}, we show the oversegmentation results of our method and the competing algorithms on vKITTI3D and S3DIS datasets. We observe that our supervized partition framework produces superpoints of adaptive sizes which closely follow hard-to-segment objects such as white boards or sidewalks.
We also notice that the embeddings learn to ignore certain form of intra-object variability of geometry and radiometry. In particular, the lamp reflections on the white boards are almost completely ignored by the embeddings. Even more interestingly, the embeddings of trees are homogeneous despite the significant variability between leafs and trunks. As a consequence, the trees are segmented into one component while the other methods produces many dubious superpoints.

\begin{table*}[ht]\centering
\begin{tabular}{ccc|cc}
\bf parameter & \bf \!\!\!shorthand &\, \bf section\,\, & \,\,\bf S3DIS & \bf vKITTI \\\hline
Local neighborhood size & $k$ & 3.1 & \multicolumn{2}{c}{20}\\
\# parameters & - &- & \multicolumn{2}{c}{13,816}\\
$\LCE$ configuration & - & 3.1 & \multicolumn{2}{c}{[32,128],[64,32,32,m]}\\
ST configuration & - & 3.1 & \multicolumn{2}{c}{[16,64],[32,16,4]}\\
Embeddings dimension & $m$ & 3.1 & \multicolumn{2}{c}{4}\\
Adjacency graph & $G$ & 3.2 & 5-nn & 5-nn + Delaunay\\
exponential edge factor & $\sigma$ & 3.2.1 & \multicolumn{2}{c}{0.5}  \\
intra-edge factor &$\tilde{\mu}$ & 3.2.3 & \multicolumn{2}{c}{5} \\
spatial influence&$\aspat$  & 3.4  & 0.2 & 0.02 \\
smallest superpoint & $n_\text{min}^{(1)}$ & 3.4 & 40 & 10 \\
epochs & - & - & \multicolumn{2}{c}{50}\\
decay event & - & - & 
\multicolumn{2}{c}{20,35,45}\\
\end{tabular}
\caption{Configuration of the embedding network for the S3DIS and vKITTI datasets.\\}
\label{tab:parameters}
\end{table*}

%.........................................................................
\section{Ablation Study}
%.........................................................................
We present an ablation study to empirically justify some of our design choices. In particular we
present \textbf{Prop-weight}, an alternative version in which the cross-partition weighting is replaced by a simple inversely-proportional weighting of the inter/intra edges. Predictably, this method gives lesser results as the edges are not weighted according to their influence in the partition. However, since the weights of the intra-edge are proportionally higher, the border precision is improved.

We replaced our choice of function $\phi$ and $\psi$ in the loss by respectively $\mid \cdot \mid$ and $-\mid \cdot \mid$, so that our loss is closer to the pairwise affinity loss used by \cite{engelmann2018} (but still structured by the graph). However, this approach wouldn't give meaningful partition as the intra-edge term conflicts with the constraint that the embeddings are constrained on the sphere. Removing this restriction leads the collapse of the embeddings around $0$. 
%We also tried to stack the $\LCE$ in layers, using or not a residual structure comparable to the one used in \cite{he2016deep} to increase their receptive fields (more details are given in the appendix). The best results were achieved with two layers: \textbf{2-Layers} and \textbf{2-Residuals}. However, we observe that when compared with $\LCE$ of a similar number of parameters, the gains are insignificant if not null. We conclude that to embed points in order to detect borders, a small receptive field with a shallow architecture is sufficient.
%
%####################

\section{Models configuration}
%**********************************

Our supervized oversegmentation model has a number of critical hyper-parameters to tune, given in \tabref{tab:parameters}. We detail here the rationale behind our choices.

\noindent
\textbf{Local neighborhood and adjacency graphs:} For both datasets, we find that setting the local neighborhood size to $20$ was enough for embeddings to successfully detect objects' borders. Combined with our lightweight structure, this results in a very low memory load overall. The adjacency graph $G$ requires more attention depending on the dataset. For the dense scans of S3DIS, the $5$-nearest neighbors adjacency structure was enough to capture the connectivity of the input clouds. For the sparse scans of vKITTI, we added Delaunay edges \cite{delaunay1934sphere} (pruned at 50 cm) such that parallel scans lines would be connected.\\

\noindent
\textbf{Networks configuration:} For the $\LCE$ and the PointNet structure in the spatial transform, we find that shallow and wide architectures works better than deeper networks. We give in \tabref{tab:parameters} the size of the linear layers, before and after the maxpool operation. Over $250,000$ points can be embedded simultaneously on 11GB RAM in the training step, while keeping track of gradients.\\

\noindent
\textbf{Intra-edge factor:}
The graph-structured contrastive loss presented in  Section 2 requires setting a weight $\mu$ determining the influence of inter-edges with respect to intra-edge. Since most edges of $G$ are intra-edges in practice, we define $\tilde{\mu}$ such that $\mu=\tilde{\mu} c$ with $c={\mid E \mid}/{\mid V \mid}$ the average connectivity of $G$. Note that $c$ can be determined directly from the construction of the adjacency graph (it is equal to $k$ in a $k$-nearest neighbor graph for example). A value of $\tilde{\mu}=1$ means that the total influence in $\ell$ of inter-edges and intra-edges are identical. Since we are interested in oversegmentation, we set $\tilde{\mu}$ to $5$ in all our experiments, but note that the network is not very sensitive to this parameter, as demonstrated experimentally: a value of $\tilde{\mu}=3$ gives a relative performance of $(-0.2,-0.6,+1.5)$ while a value of $8$ gives $(+0.1,-0.5,+1.4)$.\\

\noindent
\textbf{Regularization Strength:} The generalized minimal partition problem defined in Section 2 requires setting the regularization strength factor $\lambda$, determining the cost of  edges crossing superpoints. We remark that the $\LCE$ produces embeddings of points with an euclidean distance of at least $1$ over predicted objects' borders. Some calculus shows us that for a $\lambda\leq 1/(2 c)$, the solution $e^\star$ of (1) should predict superpoints borders at all edges whose vertices have a difference of embeddings of at least $1$ (note that there is no guarantee that the greedy $\ell_0$-cut pursuit algorithm will indeed predict a border). We use this value to define a normalized regularization strength $\tilde{\lambda}$ such that $ \lambda = \tilde{\lambda}/(4c)$, whose default value is $1$.\\
\noindent
\textbf{Regularization path:} To obtain the regularization paths in Figure 4, we first train the network with a regularization strength of $\tilde{\lambda}=1$ (see Section 2). We then compute partitions with $\tilde{\lambda}$ varying from $0.2$ to $6$ with no fine-tuning required.\\

\noindent
\textbf{Smallest superpoint:} 
To automatically select a minimal superpoint size (in number of points) appropriate to the coarseness of the segmentation, we heuristically set:
$$
n_\text{min}^{\tilde{\lambda}}=\Bra{(\max\Pa{\frac12 n_\text{min}^{(1)},n_\text{min}^{(1)} + \frac12 n_\text{min}^{(1)}\log(\tilde{\lambda})}}
$$
where $n_\text{min}^{(1)}$ is a dataset-specific minimum superpoints size for $\tilde{\lambda}=1$. For example, for $n_\text{min}^{(1)}=50$, the smallest superpoint allowed for a small regularization strength $\tilde{\lambda}= 0.2$ will be $33$, while it is $70$ for the coarse partition obtained with $\tilde{\lambda}=6$. While specific applications may require setting up this variable manually, this allowed us to produce the regularization paths in Figure 4 while only varying $\tilde{\lambda}$.\\

\noindent
\textbf{Optimization:}
Given the small size of our network, we train it for a short number of epochs (see \tabref{tab:parameters}), with decay events set at $0.7$. We use Adam optimizer \cite{kingma2014adam} with gradient clipping at $1$ \cite{goodfellow2016deep}. Training takes around 2 hours per fold on our $11$GB VRAM $1080$Ti GPU.\\

\noindent
\textbf{Mini-batches:}
For graph-based clustering, the training phase processes batches of $16$ point clouds at once, for which a subgraph of size $10\,000$ points is extracted. For the clustering-based segmentation, which is more memory intensive, and since subgraphs have to be larger to be meaningfully covered by the initial voxels, we set a batch size of $1$ and a subgraph of $100\,000$. As a consequence, we replace the batchnorm layers of the $\LCE$s by group norms with $4$ groups \cite{wu2018group}.\\

\noindent
\textbf{Augmentation:} In order to build more robust networks, we added Gaussian noise of deviation $0.03$ clamped at $0.1$ on the normalized position and color of neighborhood clouds. We also added random rotation of the input clouds for the network to learn rotation invariance. To preserve orientation information, the clouds are rotated as a whole instead of each neighborhood. This allows the spatial transform to detect change in orientation, which can be used to detect borders.

\setlength\tabcolsep{3.2pt}
%\setlength{\belowcaptionskip}{-10pt}
%\setlength{\belowcaptionskip}{0pt}
%##############
%\clearpage
%\balance
%{\small
%\bibliographystyle{icml2019}
%\bibliography{mybib}
%}

\end{document}